\documentclass[conference]{IEEEtran}
\IEEEoverridecommandlockouts
\usepackage{cite}
\usepackage{amsmath,amssymb,amsfonts}
\usepackage{algorithmic}
\usepackage{graphicx}
\usepackage{textcomp}
\usepackage{xcolor}
\def\BibTeX{{\rm B\kern-.05em{\sc i\kern-.025em b}\kern-.08em
    T\kern-.1667em\lower.7ex\hbox{E}\kern-.125emX}}

\newcommand\blfootnote[1]{%
  \begingroup
  \renewcommand\thefootnote{}\footnote{#1}%
  \addtocounter{footnote}{-1}%
  \endgroup
}

\begin{document}

\title{Toward Improving the Evaluation of Visual Attention Models: a Crowdsourcing Approach}

\author{\IEEEauthorblockN{Dario Zanca}
\IEEEauthorblockA{\textit{University of Siena} \\
Siena, Italy \\
dario.zanca@unisi.it}
\and
\IEEEauthorblockN{Stefano Melacci}
\IEEEauthorblockA{\textit{University of Siena} \\
Siena, Italy \\
mela@diism.unisi.it}
\and
\IEEEauthorblockN{Marco Gori}
\IEEEauthorblockA{\textit{University of Siena} \\
Siena, Italy \\
marco@diism.unisi.it}}

\maketitle

\begin{abstract}
Human visual attention is a complex phenomenon. A computational modeling of this phenomenon must take into account \textit{where} people look in order to evaluate which are the salient locations (spatial distribution of the fixations), \textit{when} they look in those locations to understand the temporal development of the exploration (temporal order of the fixations), and \textit{how} they move from one location to another with respect to the dynamics of the scene and the mechanics of the eyes (dynamics). 
State-of-the-art models focus on learning saliency maps from human data, a process that only takes into account the spatial component of the phenomenon and ignore its temporal and dynamical counterparts.
In this work we focus on the evaluation methodology of models of human visual attention. We underline the limits of the current metrics for saliency prediction and scanpath similarity, and we introduce a statistical measure for the evaluation of the dynamics of the simulated eye movements. While deep learning models achieve astonishing performance in saliency prediction, our analysis shows their limitations in capturing the dynamics of the process. We find that unsupervised gravitational models, despite of their simplicity, outperform all competitors. Finally, exploiting a crowd-sourcing platform, we present a study aimed at evaluating how strongly the scanpaths generated with the unsupervised gravitational models appear plausible to naive and expert human observers.
\end{abstract}


\section{Introduction}
\blfootnote{Accepted for publication at the IEEE International Joint Conference on Neural Networks (IJCNN) 2020 (DOI: TBA).

© 2020 IEEE. Personal use of this material is permitted. Permission from IEEE must be obtained for all other uses, in any current or future media, including reprinting/republishing this material for advertising or promotional purposes, creating new collective works, for resale or redistribution to servers or lists, or reuse of any copyrighted component of this work in other works.}
A huge amount of visual information constantly reaches our eyes during daily activities~\cite{koch2006much}. A visual scene typically contains much more items than the human visual system can process. Visual attention refers to a series of cognitive operations that allow us to focus on salient elements and filter out the irrelevant information~\cite{McMains2009}. The study of this process is at the crossroad of different disciplines such as neuroscience, cognitive science, computer vision, psychology. Many computational models of human attention have been developed in the last three decades (see~\cite{borji_2013,stateoftheart} for an extensive analysis of the state-of-the art), and the increasing interest in this topic is also due to a wide range of possible applications, including object detection~\cite{frintrop2006vocus}, video compression~\cite{hadizadeh2013saliency}, advertising~\cite{wolfe2011visual} or visual tracking~\cite{mahadevan2009saliency}, among others.

Nevertheless, we are still far from formalising a mechanism of attention that approximates human capabilities. Inspired by the idea of~\cite{treisman1980feature,koch1987shifts}, and following the path traced out by the seminal works of~\cite{itti1998model,harel2007graph,bruce2007attention}, state-of-the-art models focus on learning saliency from human data. This trend tacitly assumes a centralized role of the saliency map and that fixations may be eventually generated according to the \textit{Winner-Take-All} algorithm described in~\cite{koch1987shifts}. For this reason, these models are commonly evaluated with saliency metrics that take into account only the spatial component of this phenomenon, i.e. the spatial distribution of the fixations, while the temporal dynamics of the attention are not considered.  Saliency oriented models do not capture the dynamics of the mechanism but an overall statistic that tells us little about the neuroscience of visual attention. We stress out here that overclaimed conclusions should not be drawn from these attempts and more in-depth evaluation methods are necessary.
Models of scanpath that take into account the temporal order of fixations have been proposed as well, but they are often task-specific (exploration of shapes~\cite{renniger} or action recognition~\cite{sminchi}) and not easily exploitable in a free-viewing scenario. Recently, a general purpose computational description of attention as a dynamic process has been presented by~\cite{zanca2017variational}, where laws of eye movements are described in the framework of mechanics. The authors propose a mathematical formulation based on a few fundamental principles somehow connected with human attention, such as the boundedness of the retina, the curiosity towards differences in brightness, and the property of brightness invariance. 
Despite being oriented to scanpath modeling, this approach leads to impressive results in unsupervised saliency prediction (see the large comparison performed by~\cite{borji2018saliency}), while an evaluation of the quality of the predicted scanpaths has not been performed. 
Moreover, the fundamental principles mentioned above, although very general, are too local, since they do not provide a way to aggregate information from the peripheries of the visual field, and they lack a mechanism that avoids revisiting recently visited locations, which might generate unnatural trajectories when exploring the input stream. A recent approach proposes an explanation of visual attention trough gravitational models~\cite{zanca2019gravitational}. This results in an unsupervised scanpath-oriented model in which attention emerges as a dynamic process. Attention is modeled as a unitary mass subject to gravitational attraction, where the gravitational field is induced by masses associated to visual features, such as image details, motion, and, if needed, task-related information.
The output of the model is a continuous function that describes the trajectory of the focus of attention. Similarly to~\cite{zanca2017variational}, saliency can be obtained as a by-product, summing up the most visited locations.

With the aim of  improving the evaluation methodology of models of human visual attention, we underline the limits of the current metrics for scanpath similarity, and we introduce a statistical measure for the evaluation of the dynamics of the simulated eye movements. All the different approaches are tested both in saliency and scanpath prediction. Despite of their simplicity, the analysis of the results shows that gravitational models oriented to capture the dynamics of the phenomenon (instead of estimating the saliency map) outperform other approaches. Finally, with emphasis to gravitational models, we present a study of the opinions of human evaluators, collected through a crowd-sourcing platform. To the best of our knowledge, this is the first time that this type of analysis is conducted to evaluate computational models of visual attention. 

This paper is organized as follows. We review graviational models of visual attention in Section \ref{model}. An in-depth discussion on the problem of evaluating models of visual attention is presented in section~\ref{limits}. An experimental evaluation and comparisons with state-of-the-art models are presented in section~\ref{experiments}.  Mathematical formulation of the model is given in section~\ref{model}, together with results of the crowd-sourcing evaluation.

\section{Gravitational models of visual attention}\label{model}
The analysis of most of this paper is based on gravitational models of visual attention, that are recent models that have shown to yield state-of-the-art performances in unsupervised scanpath prediction \cite{zanca2019gravitational}. These models are able to generate a dynamic scanpath trajectory without the need of producing a saliency map first, thus fully relying on a differential equation that drives the focus of attention.

In order to describe the gravitational model of \cite{zanca2019gravitational}, we consider a generic stream of visual input, that is defined on the domain
$$\mathcal{D}=\mathcal{R} \times \mathcal{T}\ ,$$ 
where the subset $\mathcal{R} \subset \mathbb{R}^2$ represents the retina coordinates while $\mathcal{T} \subset  \mathbb{R}$ is the temporal domain.
The visual attention scanpath is the trajectory $a(t):\ \mathcal{T} \rightarrow \mathcal{R}$, being $t \in \mathcal{T}$ the time index.
Attention is driven by the attraction triggered by {relevant visual features} of the visual input. 
Let $f_i$ be the function associated to the activation of a visual feature, modeling the presence of a certain property in a pixel of the input stream, i.e.,
$$f_i:\mathcal{D} \to \mathbb{R}\ .$$ 
Larger values of $f_i(x,t)$ correspond with more evident presence of the visual feature in $(x,t) \in \mathcal{D}$, being $x$ the pixel coordinates.
Let us assume to have the use of a number of $f_i$'s, each of them associated to different properties of the input stream.

\begin{figure}
\centering
\includegraphics[width=0.5\textwidth]{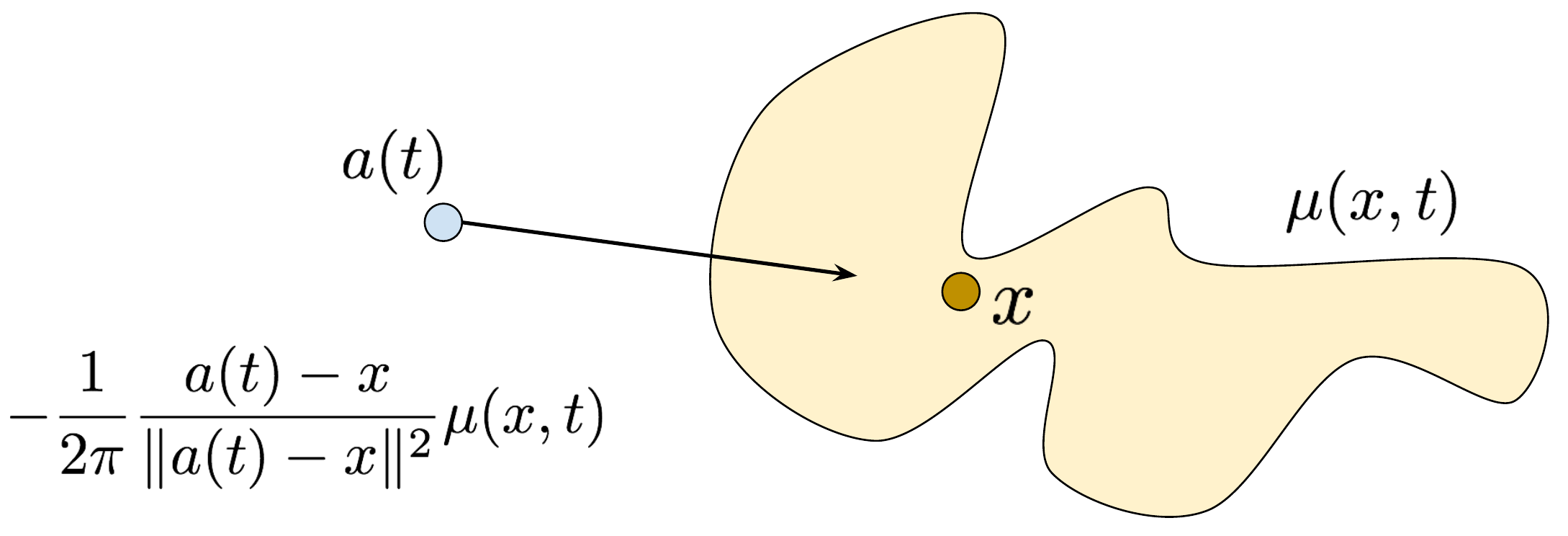}
\caption{The focus of attention represented as an elementary mass at coordinates $a(t)$ subject to a gravitational field that depends on the distributional mass $\mu$ (that is non-zero -- and not constant -- in the yellowish region). We explicitly show the attraction yielded by point $x$ (bottom-left expression).}
\label{example}
\end{figure}

Inspired by the behaviour of gravitation fields, the visual attention scanpath can be modeled as the motion of a unitary mass subject to the gravitational attraction of a distribution of masses $\mu$, associated to the visual features, $$\mu: \mathcal{D} \to \mathbb{R} \ .$$
In particular, $\mu(x,t)$ is defined as $\mu(x,t)= \sum_{i}\mu_i(x,t)$, being $\mu_i$ the mass associated to feature $f_i$, that is
$$\mu_i(x,t)= \alpha_i \lVert f_i(x,t) \rVert\ ,$$
where the norm $\Vert \cdot \Vert$ measures the strength of the activation of $f_i$, and $\alpha_i > 0$ is a customizable scaling factor.
The gravitation field $E$ \cite{feynman1965feynman} is such the attraction toward the distributional mass $\mu$ is inversely proportional to the squared distance from the focus of attention $a(t)$, and it is given by
\begin{eqnarray} 
\nonumber E(a(t),{t}) &=& - \frac{1}{2\pi} 
\int_\mathcal{R} dx \frac{a(t)-x}{\lVert a(t)-x\lVert^2} \mu(x,t) \\
 &:=& - (e * \mu)(a(t),t) \ ,
\label{overall_field}
\end{eqnarray}
where $*$ is the convolution operator and $e(z) = (2\pi)^{-1}(z) \Vert z \Vert ^{-2}$.
A sketch of this idea is reported in Fig.~\ref{example}. 
Once we are given the gravitational field, the {Newtonian differential equation} of attention are
\begin{align}
\ddot a(t) + \lambda \dot a(t) + (e \ast \mu)(a(t),{t}) = 0,
\label{Trajectory}
\end{align}
where dumping term $\lambda \dot a(t)$, with $\lambda > 0$, prevents from oscillations typical of gravitational systems and it helps to produce precise ballistic movements toward the salient target. 
Integrating Eq.~\ref{Trajectory} allows us to compute the visual attention trajectory at each time instant.\footnote{We converted the equation to a first-order system of differential equations, as commonly done, introducing auxiliary variables. Then we used the \texttt{odeint} function of the Python SciPy library, in the setting in which it automatically determines where the problem is stiff and it chooses the appropriate integration method.}

The choice of the visual features that induce the corresponding masses is determinant in modeling the behaviour of the attention system. A key property of the this model is that there are no restrictions on the categories of features one could consider. While some of the features can be pretty generic and not associated to high-level semantics of the observed input stream (e.g., variations of brightness, motion, etc.), other features could be associated to semantic categories (faces, objects, actions, etc.) that might be relevant in specific visual exploration tasks.
The features we consider in this paper are described as follows.
\begin{itemize}
	\item Let $b:\mathcal{D} \to \mathbb{R}$ be 
	  the brightness of the video, that yields the feature associated to \textit{spatial gradient of the brightness},
	$f_1 = \nabla_x b $. 
	This features carries information about edges and, generally speaking, 
	it reveals the presence of details in the input data (being it a fixed image or a video).
	\item  Let $v: \mathcal{D} \to \mathbb{R}$ be the \textit{optical flow}, that is the velocity field at any $(x,t) \in \mathcal{D}$. The feature $f_2 = v$ characterizes  moving areas in the retina. This feature only applies in the case of video streams, and we computed it using off-the-shelf implementations of the optical flow.
	\item Let $h: \mathcal{D} \to \mathbb{R}$ be the probability of the \textit{presence of a human face} at any $(x,t) \in \mathcal{D}$. The feature $f_3 = h$ is active in those areas of the retina characterized by the presence of human faces.
\end{itemize}
More features could be considered as well, by simply introducing new visual feature functions. While $f_1$ is what we constantly used in all our experiments (Section~\ref{exp}), $f_2$ and $f_3$ were only used in human evaluations, where video streams are considered too (thus enabling $f_2$) and where we also injected contribute from $f_3$, since faces are known to attract human attention in a task-independent way \cite{cerf2009faces}.

In humans,  after a reflexive shift of attention towards the source of stimulation, there is an inhibition to remain in the same location~\cite{posner1985}. This mechanism is called Inhibition Of Return (IOR). A similar mechanism is defined in the gravitational model, to prevent the trajectory to get trapped into regions of equilibrium and favour complete exploration of the scene. The dynamic of a function of inhibition $I(x,t)$ can be modeled as  
\begin{equation}
\frac{\partial I(x,t)}{\partial t} + \beta I(x,t) = \beta g(x-a(t)),
\label{IoR-Eq}
\end{equation}
where $ g(u) =  e^{- \frac{u^{2}}{2 \sigma^{2}}}$ and $0 < \beta <1$. This is directly applied to the feature masses, in order to decrease the gravitational contribution from already-visited spatial locations. As a results, the distribution of masses $\mu$ becomes
\begin{align}
\mu(x,t)=\sum_i \mu_{i}(x,t) (1-I(x,t)) \ .
\label{OverallMass}
\end{align}

\section{Evaluating visual attention dynamics}\label{limits}

A number of papers in the last three decades have compared models of visual attention across different datasets~\cite{Judd,kootstra,bruce2007attention,cat2000} and saliency metrics, such as the distribution-based  Kullback-Leibler divergence (KL)~\cite{kullback1951information}, the location-based Area Under the Curve (AUC)~\cite{riche2013saliency}, and the Normalized Scanpath Saliency (NSS)~\cite{peters2005components}.
Different metrics give different importance to the presence of false positives and false negatives in the predicted saliency map, when compared to ground truth human fixations. Moreover, they can be differently affected by systematic viewing biases, such as the center bias~\cite{bylinskii2019different}. The problem of evaluating saliency models has been deeply studied and a set of qualitative and quantitative properties of saliency metrics has been investigated over years~\cite{wilming2011measures,borji_2013,bylinskii2019different}.

\begin{figure*}[ht]
	\begin{center}
\begin{tabular}{ccc}
\includegraphics[width=.14\linewidth]{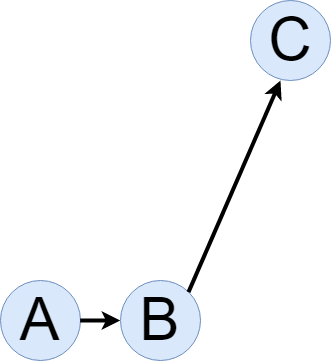} & 
\includegraphics[width=.14\linewidth]{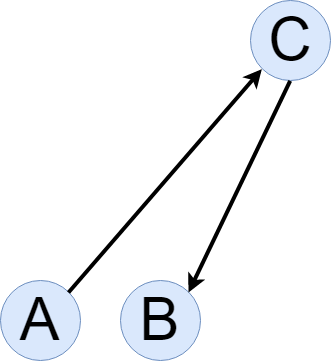} & 
\includegraphics[width=.14\linewidth]{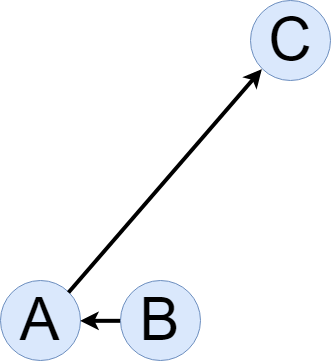} \\ 
\\
\footnotesize
\textsc{Human} & 
\footnotesize $\quad$ \textsc{Synthetic scanpath} $1$ $\quad$  & 
\footnotesize $\quad$  \textsc{Synthetic scanpath} $2$ $\quad$ 
\end{tabular}
\end{center}
\caption{\textbf{Example of scanpaths.} The three scanpaths visit exactly the same three spatial locations $A$, $B$ and $C$, but with a different temporal order.}
\label{scanpath_comparison}
\end{figure*} 

In the computer vision literature, it is less frequent to find studies on the problem of evaluating computational models of visual attention taking into account  the temporal order of the fixations, in addition to the widely considered spatial distribution of such fixations, i.e., the saliency map. There exists a number of tools for measuring the similarity between human and simulated \textit{visual scanpaths}\footnote{A visual scanpath is defined as an ordered sequence of fixations.}. 
Some authors use the string-edit (Levenshtein) distance (SE)~\cite{Jurafsky,Brandt,Foulsham}, where the visual input is divided into $n \times m$ regions, uniquely labeled with a character. Then, each scanpath can be associated with a string, taking the ordered sequence of labels of the regions in which the fixations fall. The distance between strings is an indicator of the distance between the corresponding scanpaths. In~\cite{Choi}, the string-edit distance has been shown to be a robust metric with respect to changes in the number of considered regions. In~\cite{borji_2013}, a number of saliency models are used to generate scanpaths, and their performances are evaluated with a slightly modified version of the SE. Other authors proposed a scaled time-delay embedding (STDE)~\cite{wang2011simulating,FixaTons} measure of similarity, which derives from a popular metric for a quantitative comparison of stochastic and dynamic trajectories of varied lengths, in the filed of physics. 

However, the widely used saliency and scanpath metrics do not evaluate some important properties on the dynamics of the exploration, that we emphasize in the following example.
Let $A\rightarrow B\rightarrow C$ be a true (human) scanpath across three spatial locations $A$, $B$, $C$, and let $A\rightarrow C\rightarrow B$ and $B\rightarrow A\rightarrow C$ be two synthetic (simulated) scanpaths generated with two different models of visual attention, as shown in Fig.~\ref{scanpath_comparison}. Both the models visit exactly the same three spatial locations that are visited by the human scanpath, but the three scanpaths differ in the order in which these locations are visited. Since the spatial distribution of the fixation is identical, a saliency metric will indicate a perfect saliency prediction in both the synthetic cases. Differently, visual-scanpath-oriented metrics, such as SE, will capture some differences. As a matter of fact, the string-edit distance between each of the two synthetic scanpaths and the human scanpath is equal to $1$ (only an \textit{exchange} operation in the string is needed). However, we would have reason to say that the \textit{synthetic scanpath} $2$ of Fig.~\ref{scanpath_comparison} is better than the \textit{synthetic scanpath} $1$ since it yields an initial short saccade, similarly to what happens in the human case. Differently, the \textit{synthetic scanpath} $1$ is only based on long saccades, making it less closer to the human scanpath. 

In this specific case, it may be useful to study statistical quantities related to the dynamics of the phenomenon under examination. In particular, the distribution of saccade amplitudes provide statistical information that is not captured by the aforementioned popular metrics. This statistical quantity has been previously used in evaluating the quality of computational models of attention~\cite{lemeur_liu,boccignone2004modelling}, in a context in which human exploration biases were added to the model. 
We propose to evaluate artificially generated scanpaths not only with classic metrics, but also with the KL divergence between the distributions of amplitudes of human saccades and of artificially generated ones. 

Despite introducing some precious information, the proposed evaluation methodology is still not enough. A number of dynamic patterns of visual exploration can characterize the human scanpath. Some may concern the mechanics of the eyes, others the visual patterns of the scene, or other high-level semantics. Furthermore, there exists a wide variability among human subjects. 
While the definition of an all-inclusive metric is probably not possible, we can evaluate how strongly a synthetic scanpath is \textit{plausible} (i.e. "human-like" or "natural") by collecting feedbacks from uninformed observers which may be sensible to uncommon behaviours, unnatural vibrations, meaningless explorations. For this reason, we propose to complement the experimental analysis based on metrics with a crowd-sourcing-based evaluation, in which human evaluators are asked to tag scanpaths as "human-like" or "artificial". A statistical study of the collected evaluator opinions provides an indication on the qualitative plausibility of the output of a computational model.

\section{Experimental evaluation and analysis}\label{experiments}
\label{exp}



In what follows, we evaluate a number of different visual attention models following all the strategies of Section~\ref{limits}. A huge number of models are present in the literature. They have been selected in this work among the most representative of their typology. In Section~\ref{x} we briefly describe each of the selected models of visual attention. In Section~\ref{a} we evaluate the models in the tasks of saliency and scanpath prediction. Saccade amplitude statistics are compared to human statistics in Section~\ref{b}. Crowd-sourcing evaluation is performed for the case of gravitational models in section~\ref{c}.

    \subsection{State-of-the-art models of human visual attention}\label{x}
    
    The procedure described in \cite{koch1987shifts} is used to generate fixations from the selected saliency models~\cite{itti1998model,sam,kummerer2016deepgaze}. 
    \begin{itemize}
        \item \textit{SAM}~\cite{sam} and \textit{Deep Gaze II}~\cite{kummerer2016deepgaze} are the best supervised models in saliency prediction, according to the MIT Saliency Benchmark~\cite{mit-saliency-benchmark}, for the CAT2000 and MIT300 datasets respectively. Both models are based on deep learning methods and learn the salience directly from the data.
        \item \textit{Eymol}~\cite{zanca2017variational} is a scanpath-oriented unsupervised model, providing outstanding results in unsupervised saliency prediction (see~\cite{borji2018saliency}). 
        \item Gravitational models~\cite{zanca2019gravitational} define an unsupervised scanpath-oriented model in which attention emerges as a dynamic process, as described in Section \ref{model}.
        \item \textit{Itti}~\cite{itti1998model} is an unsupervised saliency model. None of the original papers evaluate the model in the task of scanpath prediction. For all experiments, we used the code provided by the authors in their public repositories.
    \end{itemize}

	\subsection{Saliency and scanpath prediction}
	\label{a}
	Our first analysis consists in benchmarking selected models using commonly used image datasets, focussing on the tasks of ($i.$) scanpath prediction and of ($ii.$) saliency prediction. In particular, the datasets used for the scanpath prediction are MIT1003~\cite{Judd}, SIENA12~\cite{FixaTons}, TORONTO~\cite{bruce2007attention}, KOOTSRA~\cite{kootstra}, while we used the well established CAT2000~\cite{cat2000} dataset for the saliency prediction task. The first $4$ datasets contain a total of $1234$ images, belonging to a wide range of different semantic categories. The resolution of the images varies from $681\times511$ to $1024\times768$ px. The CAT2000 test dataset contains $2000$ images from $20$ different categories and the resolution of the images is $1920\times1080$ px. 
    Table~\ref{tab:saliency_and_scanpath_scores} shows the results of a massive quantitative analysis on a merged collection of the aforementioned datasets of human fixations, comparing state-of-the-art approaches of visual attention.
	
	\begin{table*}[t]
		\begin{center}
			\caption{\textbf{Saliency and Scanpath prediction scores.} Larger AUC/NSS and STDE  scores are preferable, while smaller string-edit distance score correspond with better results.}
			\label{tab:saliency_and_scanpath_scores}
			\begin{tabular}{|lc|cc|cc|}
				\hline
				& &\multicolumn{2}{c|}{\textbf{Saliency prediction}} &  \multicolumn{2}{c|}{\textbf{Scanpath prediction}} \\
				\textbf{Model}&\textbf{Supervised}&{AUC}&{NSS}&{String-Edit}&{STDE}\\
				\hline\hline			
				Gravitational model&No&0.84&1.57&\textbf{7.34}&\textbf{0.81}\\				
				\hline
				Eymol&No&0.83&1.78&7.94&0.74\\
				\hline	
				SAM &Yes&\textbf{0.88}&\textbf{2.38}&8.02&0.77\\
				\hline	
				Deep Gaze II&Yes&0.77&1.16&8.17&0.72\\
				\hline		
				Itti &No&0.77&1.06&8.15&0.70\\	
				\hline		
			\end{tabular}
			\end{center}
	\end{table*}

	\begin{table*}[ht]
		\begin{center}
			
			\caption{\textbf{Crowd-sourcing evaluation statistics.} We report the the average fraction of videos that were correctly labeled (either as human or non-human). Standard deviation is in brackets.}
			\label{tab:crowd_sourcing_stats}
			
			\begin{tabular}{|c|c|c|c|c|}
				\hline
				\textbf{Overall} & \textbf{Expert evaluators} & \textbf{Naive evaluators} & \footnotesize \textbf{Human videos labeled as human} &\footnotesize  \textbf{Synthetic videos labeled as human} \\
				\hline
				0.53 (0.10)
				& 0.55 (0.11) & 0.50 (0.09) & 0.53 (0.17) & 0.46 (0.18) \\
				\hline
			\end{tabular}
		\end{center}
	\end{table*}

	The results clearly show that supervised deep learning models yield better results than scanpath oriented models in the task of saliency prediction\footnote{We calculated saliency scores for the model Deep Gaze II on the training set of CAT2000, since authors did not submit their model to the MIT Saliency Team~\cite{mit-saliency-benchmark} for the test evaluation.}, but they lack in capturing the time dynamics, and gravitational models have the best score in the scanpath prediction task.

	\begin{figure}[ht]
		\begin{center}
			\includegraphics[width=.99\linewidth]{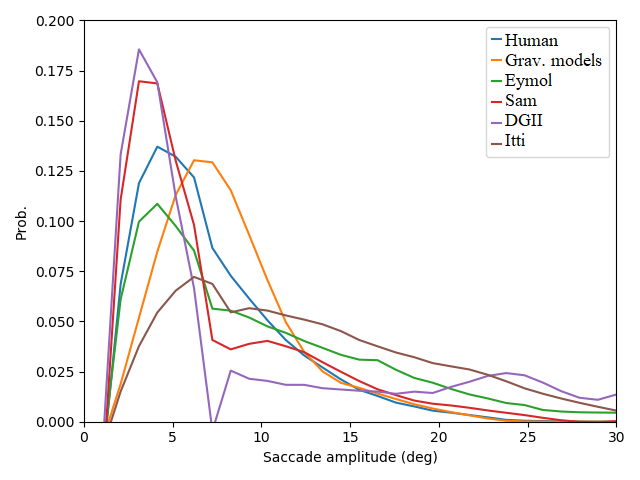}
			\caption{\textbf{Saccade amplitude distributions.} The human saccade amplitude distribution (blue) is compared with saccade distributions of the scanpaths generated with different artificial models. Data are collected in a collection of datasets composed by MIT1003~\cite{Judd}, SIENA12~\cite{FixaTons}, TORONTO~\cite{bruce2007attention} and KOOTSRA~\cite{kootstra}. Best viewed in color.}
			\label{fig:saccade_amplitude}
		\end{center}
	\end{figure} 
	
	This discrepancy was anticipated by the analysis of the metrics made in the previous section. If models based on deep learning show a surprising ability to learn associations between visual features and salience, they fail to capture the dynamics of the process. In other words, the two alternatives excel in modeling two different aspects: one related to "where" humans look, the other related to "when" or in what order they do it.

	\subsection{Saccade amplitude analysis}
	\label{b}
	This analysis, instead, wants to assess how good the models are at predicting "how" people shift attention from one location to another.
	Saliency and scanpath metrics alone cannot provide a comprehensive tool for the evaluation of visual attention models, since some aspects related to dynamics still are not captured by those metrics.
	
	Here we compare the distribution of human saccade amplitude together with the distribution generated from the simulations of the models under examination. Results are summarized in Fig.~\ref{fig:saccade_amplitude}. The plot of gravitational models is the closest to the human one, and this is further confirmed by the results in Table~\ref{tab:kl_div_sac_ampl}, that show the KL-divergence between the distribution of the saccade amplitude of the artificial attention models and that of the human scanpaths. Please note that the KL-divergence is asymmetric and for this reason the human data are taken as reference set for all the ratings in Table~\ref{tab:kl_div_sac_ampl}. Also the \textit{Eymol} model~\cite{zanca2017variational} produces competitive results. One of the motivations behind the results is that we noticed that scanpath-oriented models  favour short saccades, incorporating a principle of proximity preference which is also observed in humans~\cite{itti1998model,koch1987shifts,koch1984selecting}. 
	
	\begin{table}[h]
		\begin{center}
			
			\caption{\textbf{Saccade amplitude analysis.} KL divergence of the distribution of saccade amplitude of the scanpath generated by the models to the distribution obtained from human observers in the same images from the datasets MIT1003~\cite{Judd}, SIENA12~\cite{FixaTons}, TORONTO~\cite{bruce2007attention} and KOOTSRA~\cite{kootstra}. We compared the same models of Table~\ref{tab:saliency_and_scanpath_scores}.}
			 \label{tab:kl_div_sac_ampl}
			 \begin{tabular}{|c|c|c|c|c|}
				\hline
				\textbf{Grav. models} & \textbf{Eymol} & \textbf{Sam} & \textbf{Deep Gaze II} & \textbf{Itti} \\
				\hline
				\textbf{0.27 }& 0.46 & 1.07 & 1.44 & 2.11 \\
				\hline
			\end{tabular}
		\end{center}
	\end{table}
	
		\subsection{Crowd-sourcing evaluation}
	\label{c}
	
	\begin{figure*}		
		\begin{tabular}{ccc}
			\includegraphics[width=.31\linewidth]{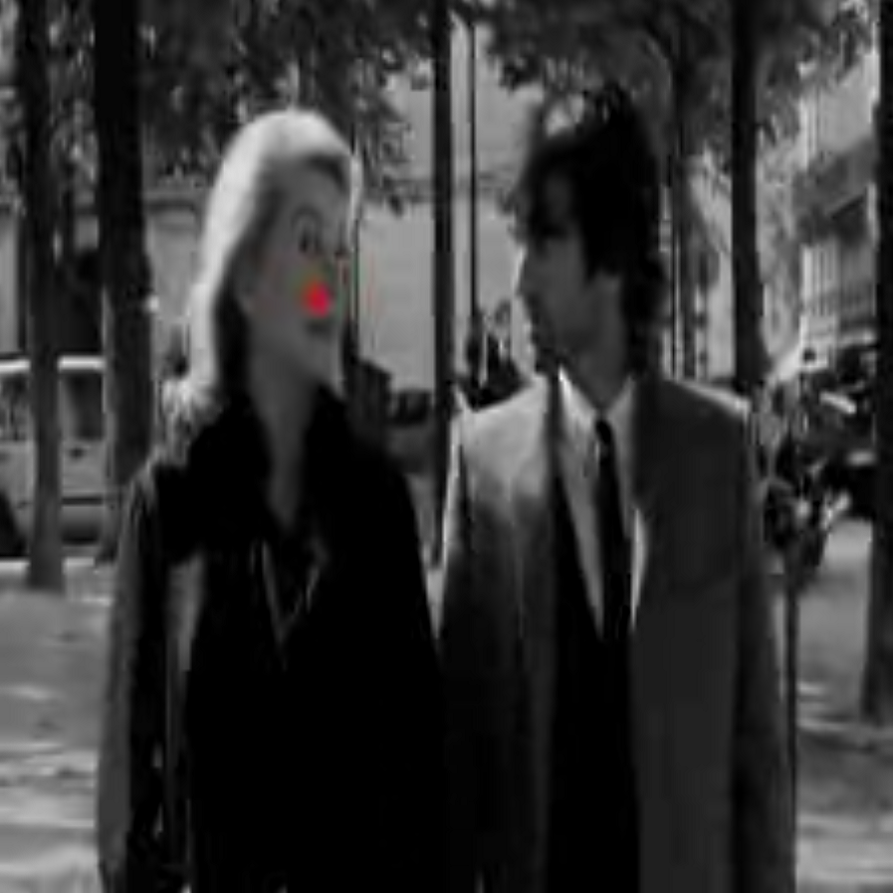} & 
			\includegraphics[width=.31\linewidth]{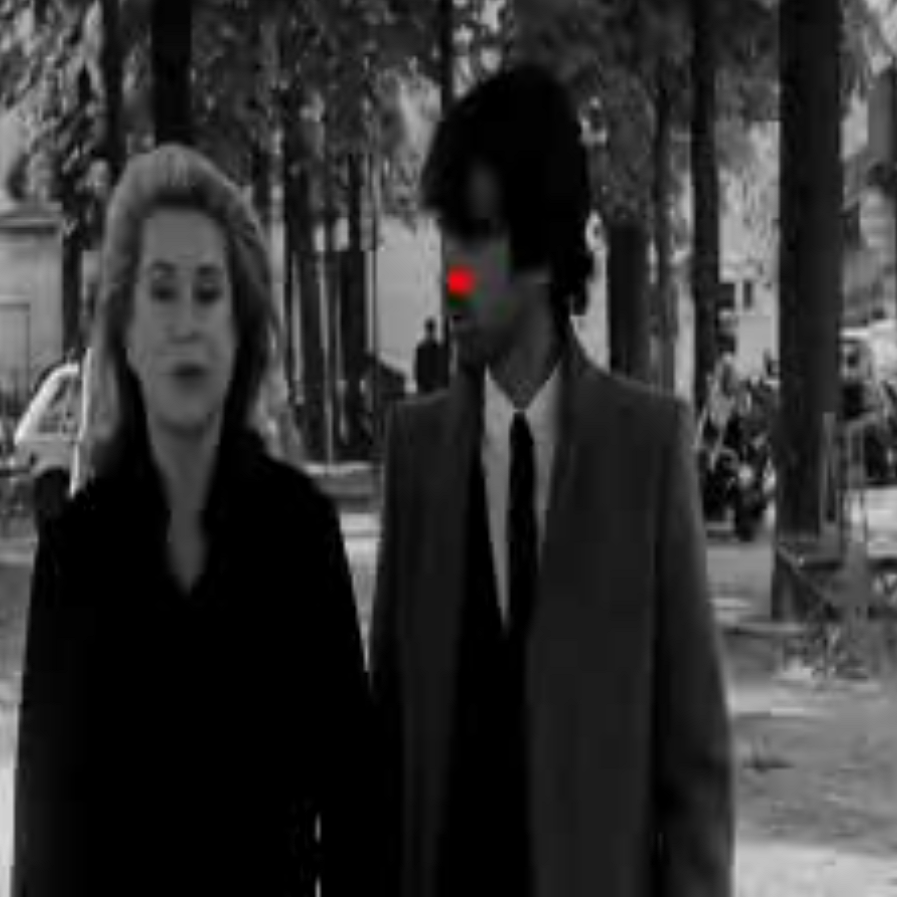} & 
			\includegraphics[width=.31\linewidth]{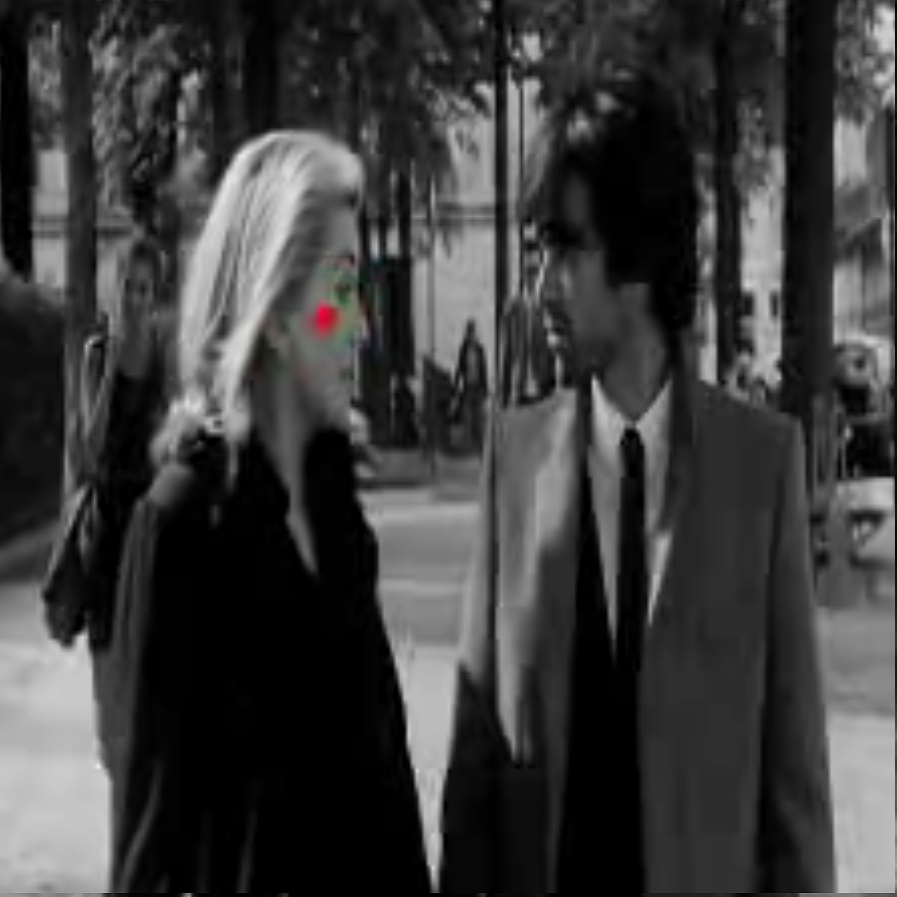}
			\\
			(a1)&(a2)&(a3) \\ 
			\includegraphics[width=.31\linewidth]{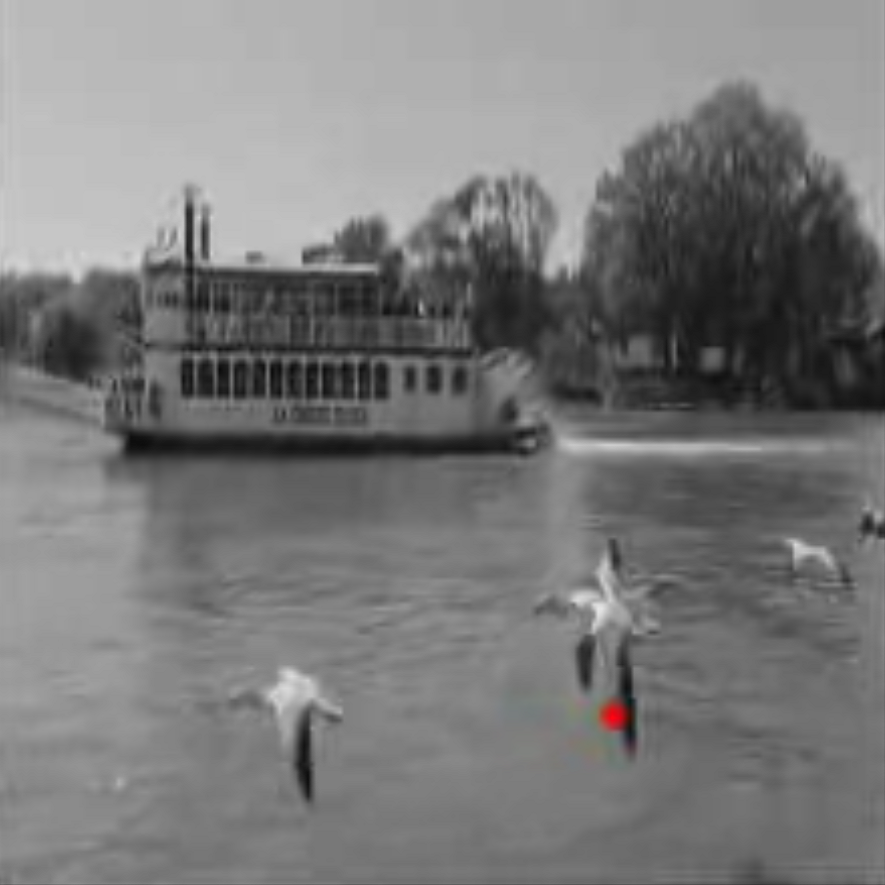} & 
			\includegraphics[width=.31\linewidth]{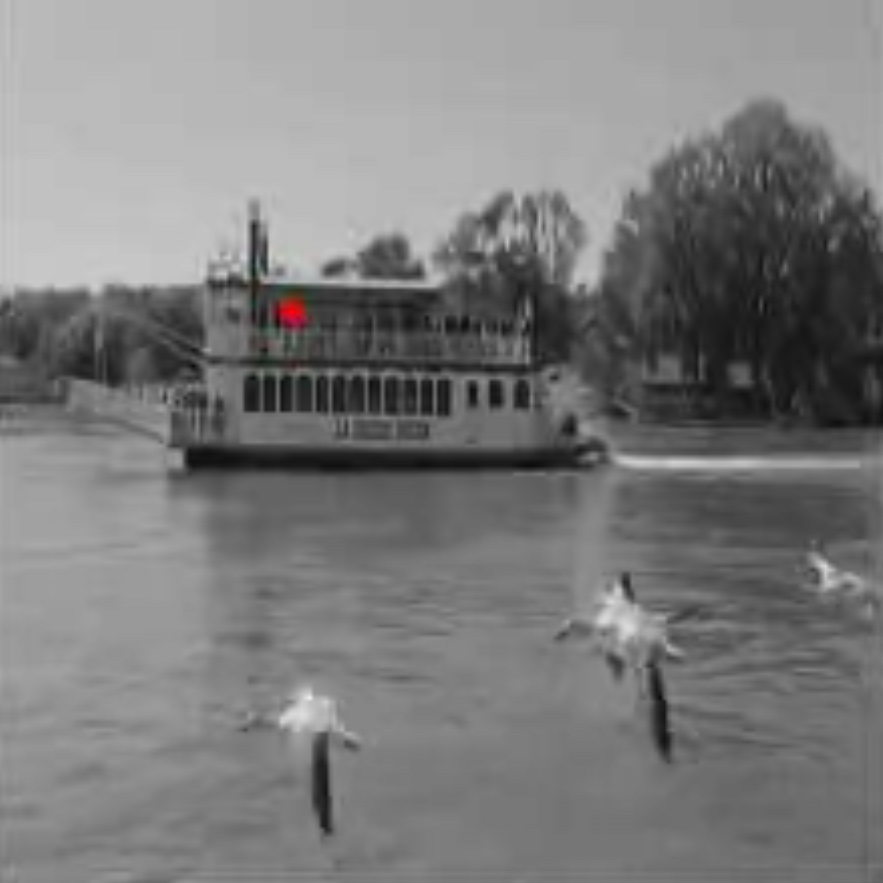} & 
			\includegraphics[width=.31\linewidth]{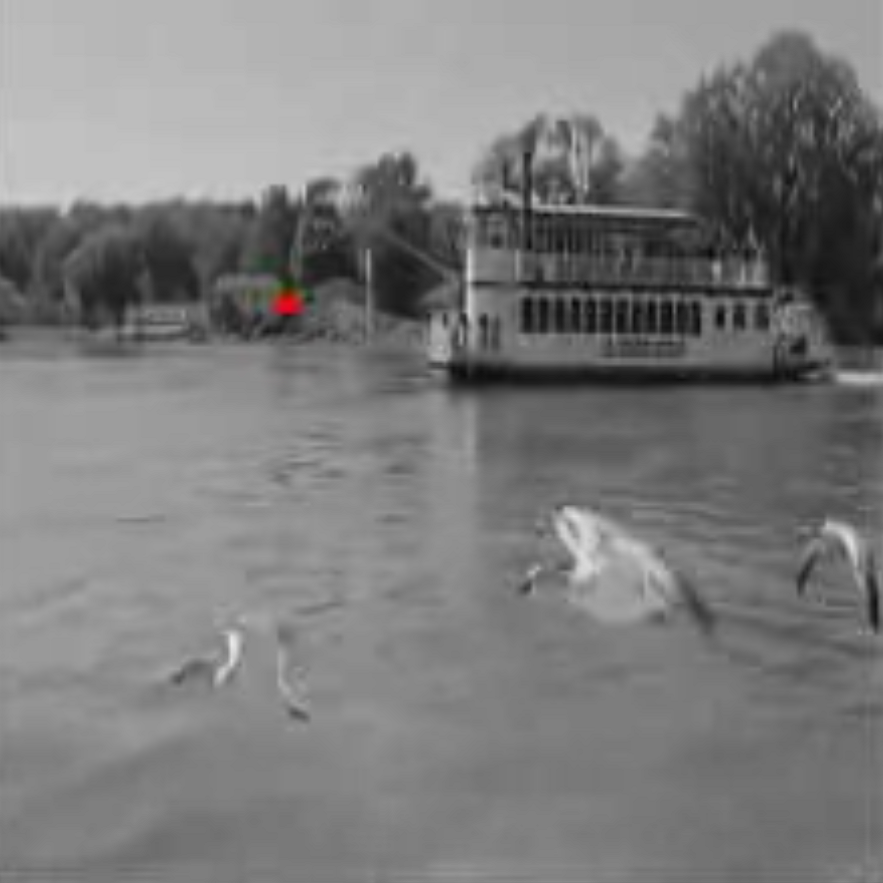}
			\\
			(b1)&(b2)&(b3)
		\end{tabular}
		\caption{\textbf{Screen-shots of scanpath presentations.} The gaze position is represented with a red filled circle in the correspondent position. Screen-shots are taken at different time steps. Best viewed in color. }
		\label{fig:redMarker}
	\end{figure*}

	We setup a crowd-sourcing evaluation procedure for testing the best performing model in scanpath predictions, i.e. the gravitational models. To this end, we used a collection of 60 videos from the {COUTROT Dataset 1}~\cite{coutrot} and 60 static images randomly sampled from MIT1003~\cite{Judd}, that are publicly available datasets of human fixations. Videos include one or several moving objects, landscapes, and scenes of people having a conversation (see supplementary material). The resolution of the video frames is $720\times 576$ px, and the average duration of each clip is $17$ seconds. Static images size varies from $405$ to $1024$ px, and they include landscape and portrait. The duration of the scanpaths in the case of static images was set to $5$ seconds. 
	
	The participants in the crowd-sourcing are presented 20 random videos of scanpaths from the aforementioned collection, in which the the gaze position is marked by a red circle, as shown in Fig.~\ref{fig:redMarker}. 
	Out of them, 10 videos are about human scanpaths, while the other 10 are about synthetic scanpaths generated with the model of Section \ref{model}. 
	Subjects are asked to evaluate each scanpath, classifying it as human or synthetic, and they provide their feedback by means of a web platform that we developed to the purpose of this evaluation. Subjects are asked some personal information about their level of education and their level of knowledge on eye movements (from 1 to 5) before starting the test. We invited 35 different subjects to participare to the crowd-sourcing, almost evenly distributed between experts on eye movements and not-experts (``naive''). 

	The statistics we collected are reported in Table~\ref{tab:crowd_sourcing_stats}.
	Results shows that the accuracy in recognizing synthetic scanpaths is close to the accuracy in recognizing human scanpaths. It is important to remark that since subjects were explicitly asked to distinguish human videos from the simulated ones, they had a natural tendency of assigning the label ``human'' only to a portion of the videos, that we found to be 49.4\% (+/- 13.7\%) of the observed videos.
	 The overall accuracy of the subjects (53\%) is very close to the random policy (50\%). This means that there are few elements that allow the observers to distinguish the human scanpaths from the synthetic ones. The expert evaluators (self-evaluated level of knowledge about eye movement between 3 and 5) have reached a score that is slightly larger than that of the naive observers (eye movement knowledge between 1 and 2). In this sense, we conclude that many aspects of the motion dynamics have been captured by the gravitational model (Section \ref{model}), as motion artefacts are normally easily perceived by experts in the field. The last two columns of Table~\ref{tab:crowd_sourcing_stats} confirms that the evaluators were in strong difficulties in discriminating human scanpaths by the artificial ones.
	 	
	In order to evaluate the agreement between annotators, we used the Fleiss' kappa~\cite{fleiss1971measuring},
	$$\kappa = \dfrac{\left(\dfrac{1}{N} \sum_{i=1}^N P_i\right) - \bar{P_e}}{1 - \bar{P_e}} \text{,}$$
	where
	$$P_i = \dfrac{\sum_{j \in \{1,2\}} n_{ij}(n_{ij} -1)}{n(n-1)},$$
	$N$ is the number of videos, $n_j$ is the number of annotators who assigned the clip  to the $j$-th category (Human or Synthetic), and $n$ is the total number of annotators. The term $\bar{P_e}$ gives the degree of agreement that is attainable by chance.
	{The quantity $P_i$ corresponds to the extent to which annotators agree on the $i$-th clip,} that is the number of pairs of evaluators that are in agreement, relative to the number of all possible evaluator pairs. Values of $\kappa$ close to 1 express complete agreement among annotators, while value of $\kappa$ lower then 0 indicate poor agreement. Analysis show a slight agreement among annotators $\kappa = 0.15$, while there is fair agreement in the case of expert annotators ($\kappa_{exp} = 0.2$, against $\kappa_{naive} = 0.09$ of the naive annotators).	Fleiss' kappa values are very similar in the case of human ($\kappa_H =  0.17$) and synthetic ($\kappa_S = 0.14$) scanpaths annotations.


\section{Conclusions and Future Work}
In this paper we presented a comparison between a selection of state-of-the-art saliency and scanpath oriented models of human visual attention. Experimental results show that the approaches that postulate the central role of saliency maps are not effective as a computational description of human visual attention as a dynamic process. Scanpath oriented models overcome saliency based approaches, despite their simplicity. In particular, gravitational models show the best results.
Great attention has been directed to the problem of correctly evaluating attention models, taking into account all the fundamental components: spatial distribution of fixations (saliency), temporal order of fixations (scanpath prediction) and movement dynamics. We have shown how certain dynamics can be captured by other statistics such as the study of saccade amplitude. Gravitational models generated saccades statistics very similar to the human ones, even if it has not been explicitly modeled for that.
For this reason we further investigated this approach with a study of the data collected with a crowd-sourcing platform. Analysis of participants opinions show that gravitational models' generated scanpaths appear plausible and are not easily distinguishable from the human ones, particularly in the case of naive annotators. 
We wish that this evaluation methodology will be applied to evaluate the attention models in a broad way from now on, making results more readable, fair and reliable, comparing to the well-established saliency benchmarks.

\bibliographystyle{IEEEtran}
\bibliography{references}

\begin{thebibliography}{10}
\providecommand{\url}[1]{#1}
\csname url@samestyle\endcsname
\providecommand{\newblock}{\relax}
\providecommand{\bibinfo}[2]{#2}
\providecommand{\BIBentrySTDinterwordspacing}{\spaceskip=0pt\relax}
\providecommand{\BIBentryALTinterwordstretchfactor}{4}
\providecommand{\BIBentryALTinterwordspacing}{\spaceskip=\fontdimen2\font plus
\BIBentryALTinterwordstretchfactor\fontdimen3\font minus
  \fontdimen4\font\relax}
\providecommand{\BIBforeignlanguage}[2]{{%
\expandafter\ifx\csname l@#1\endcsname\relax
\typeout{** WARNING: IEEEtran.bst: No hyphenation pattern has been}%
\typeout{** loaded for the language `#1'. Using the pattern for}%
\typeout{** the default language instead.}%
\else
\language=\csname l@#1\endcsname
\fi
#2}}
\providecommand{\BIBdecl}{\relax}
\BIBdecl

\bibitem{koch2006much}
K.~Koch, J.~McLean, R.~Segev, M.~A. Freed, M.~J. Berry~II, V.~Balasubramanian,
  and P.~Sterling, ``How much the eye tells the brain,'' \emph{Current
  Biology}, vol.~16, no.~14, pp. 1428--1434, 2006.

\bibitem{McMains2009}
S.~A. McMains and S.~Kastner, \emph{Visual Attention}.\hskip 1em plus 0.5em
  minus 0.4em\relax Berlin, Heidelberg: Springer Berlin Heidelberg, 2009, pp.
  4296--4302.

\bibitem{borji_2013}
A.~Borji, D.~N. Sihite, and L.~Itti, ``Quantitative analysis of human-model
  agreement in visual saliency modeling: A comparative study,'' \emph{IEEE
  Transactions on Image Processing}, vol.~22, no.~1, pp. 55--69, 2013.

\bibitem{stateoftheart}
A.~Borji and L.~Itti, ``State-of-the-art in visual attention modeling,''
  \emph{IEEE transactions on pattern analysis and machine intelligence},
  vol.~35, no.~1, pp. 185--207, 2013.

\bibitem{frintrop2006vocus}
S.~Frintrop, \emph{VOCUS: A visual attention system for object detection and
  goal-directed search}.\hskip 1em plus 0.5em minus 0.4em\relax Springer, 2006,
  vol. 3899.

\bibitem{hadizadeh2013saliency}
H.~Hadizadeh and I.~V. Baji{\'c}, ``Saliency-aware video compression,''
  \emph{IEEE Transactions on Image Processing}, vol.~23, no.~1, pp. 19--33,
  2013.

\bibitem{wolfe2011visual}
J.~M. Wolfe, G.~A. Alvarez, R.~Rosenholtz, Y.~I. Kuzmova, and A.~M. Sherman,
  ``Visual search for arbitrary objects in real scenes,'' \emph{Attention,
  Perception, \& Psychophysics}, vol.~73, no.~6, p. 1650, 2011.

\bibitem{mahadevan2009saliency}
V.~Mahadevan and N.~Vasconcelos, ``Saliency-based discriminant tracking,'' in
  \emph{2009 IEEE conference on computer vision and pattern recognition}.\hskip
  1em plus 0.5em minus 0.4em\relax IEEE, 2009, pp. 1007--1013.

\bibitem{treisman1980feature}
A.~M. Treisman and G.~Gelade, ``A feature-integration theory of attention,''
  \emph{Cognitive psychology}, vol.~12, no.~1, pp. 97--136, 1980.

\bibitem{koch1987shifts}
C.~Koch and S.~Ullman, ``Shifts in selective visual attention: towards the
  underlying neural circuitry,'' in \emph{Matters of intelligence}.\hskip 1em
  plus 0.5em minus 0.4em\relax Springer, 1987, pp. 115--141.

\bibitem{itti1998model}
L.~Itti, C.~Koch, and E.~Niebur, ``A model of saliency-based visual attention
  for rapid scene analysis,'' \emph{IEEE Transactions on Pattern Analysis \&
  Machine Intelligence}, no.~11, pp. 1254--1259, 1998.

\bibitem{harel2007graph}
J.~Harel, C.~Koch, and P.~Perona, ``Graph-based visual saliency,'' in
  \emph{Advances in neural information processing systems}, 2007, pp. 545--552.

\bibitem{bruce2007attention}
N.~Bruce and J.~Tsotsos, ``Attention based on information maximization,''
  \emph{Journal of Vision}, vol.~7, no.~9, pp. 950--950, 2007.

\bibitem{renniger}
L.~W. Renninger, J.~M. Coughlan, P.~Verghese, and J.~Malik, ``An information
  maximization model of eye movements,'' in \emph{Advances in neural
  information processing systems}, 2005, pp. 1121--1128.

\bibitem{sminchi}
S.~Mathe and C.~Sminchisescu, ``Action from still image dataset and inverse
  optimal control to learn task specific visual scanpaths,'' in \emph{Advances
  in neural information processing systems}, 2013, pp. 1923--1931.

\bibitem{zanca2017variational}
D.~Zanca and M.~Gori, ``Variational laws of visual attention for dynamic
  scenes,'' in \emph{Advances in Neural Information Processing Systems}, 2017,
  pp. 3823--3832.

\bibitem{borji2018saliency}
A.~Borji, ``Saliency prediction in the deep learning era: An empirical
  investigation,'' \emph{arXiv preprint arXiv:1810.03716}, 2018.

\bibitem{zanca2019gravitational}
D.~Zanca, S.~Melacci, and M.~Gori, ``Gravitational laws of focus of
  attention,'' \emph{IEEE transactions on pattern analysis and machine
  intelligence}, 2019.

\bibitem{feynman1965feynman}
R.~P. Feynman, R.~B. Leighton, and M.~Sands, ``The feynman lectures on physics;
  vol. i,'' \emph{American Journal of Physics}, vol.~33, no.~9, pp. 750--752,
  1965.

\bibitem{cerf2009faces}
M.~Cerf, E.~P. Frady, and C.~Koch, ``Faces and text attract gaze independent of
  the task: Experimental data and computer model,'' \emph{Journal of vision},
  vol.~9, no.~12, pp. 10--10, 2009.

\bibitem{posner1985}
M.~I. Posner, R.~D. Rafal, L.~S. Choate, and J.~Vaughan, ``Inhibition of
  return: Neural basis and function,'' \emph{Cognitive neuropsychology},
  vol.~2, no.~3, pp. 211--228, 1985.

\bibitem{Judd}
T.~Judd, K.~Ehinger, F.~Durand, and A.~Torralba, ``Learning to predict where
  humans look,'' pp. 2106--2113, 2009.

\bibitem{kootstra}
G.~Kootstra, B.~de~Boer, and L.~R. Schomaker, ``Predicting eye fixations on
  complex visual stimuli using local symmetry,'' \emph{Cognitive computation},
  vol.~3, no.~1, pp. 223--240, 2011.

\bibitem{cat2000}
A.~Borji and L.~Itti, ``Cat2000: A large scale fixation dataset for boosting
  saliency research,'' \emph{ArXiv preprint, arXiv:1505.03581}, 2015.

\bibitem{kullback1951information}
S.~Kullback and R.~A. Leibler, ``On information and sufficiency,'' \emph{The
  annals of mathematical statistics}, vol.~22, no.~1, pp. 79--86, 1951.

\bibitem{riche2013saliency}
N.~Riche, M.~Duvinage, M.~Mancas, B.~Gosselin, and T.~Dutoit, ``Saliency and
  human fixations: State-of-the-art and study of comparison metrics,'' in
  \emph{Proceedings of the IEEE international conference on computer vision},
  2013, pp. 1153--1160.

\bibitem{peters2005components}
R.~J. Peters, A.~Iyer, L.~Itti, and C.~Koch, ``Components of bottom-up gaze
  allocation in natural images,'' \emph{Vision research}, vol.~45, no.~18, pp.
  2397--2416, 2005.

\bibitem{bylinskii2019different}
Z.~Bylinskii, T.~Judd, A.~Oliva, A.~Torralba, and F.~Durand, ``What do
  different evaluation metrics tell us about saliency models?'' \emph{IEEE
  transactions on pattern analysis and machine intelligence}, vol.~41, no.~3,
  pp. 740--757, 2019.

\bibitem{wilming2011measures}
N.~Wilming, T.~Betz, T.~C. Kietzmann, and P.~K{\"o}nig, ``Measures and limits
  of models of fixation selection,'' \emph{PloS one}, vol.~6, no.~9, p. e24038,
  2011.

\bibitem{Jurafsky}
D.~Jurafsky and J.~H. Martin, \emph{Speech and language processing}.\hskip 1em
  plus 0.5em minus 0.4em\relax Pearson London, 2014, vol.~3.

\bibitem{Brandt}
S.~A. Brandt and L.~W. Stark, ``Spontaneous eye movements during visual imagery
  reflect the content of the visual scene,'' \emph{Journal of cognitive
  neuroscience}, vol.~9, no.~1, pp. 27--38, 1997.

\bibitem{Foulsham}
T.~Foulsham and G.~Underwood, ``What can saliency models predict about eye
  movements? spatial and sequential aspects of fixations during encoding and
  recognition,'' \emph{Journal of vision}, vol.~8, no.~2, pp. 6--6, 2008.

\bibitem{Choi}
Y.~S. Choi, A.~D. Mosley, and L.~W. Stark, ``String editing analysis of human
  visual search.'' \emph{Optometry and vision science: official publication of
  the American Academy of Optometry}, vol.~72, no.~7, pp. 439--451, 1995.

\bibitem{wang2011simulating}
W.~Wang, C.~Chen, Y.~Wang, T.~Jiang, F.~Fang, and Y.~Yao, ``Simulating human
  saccadic scanpaths on natural images,'' in \emph{CVPR 2011}.\hskip 1em plus
  0.5em minus 0.4em\relax IEEE, 2011, pp. 441--448.

\bibitem{FixaTons}
D.~Zanca, V.~Serchi, P.~Piu, F.~Rosini, and A.~Rufa, ``Fixatons: A collection
  of human fixations datasets and metrics for scanpath similarity,''
  \emph{ArXiv preprint, arXiv:1802.02534}, 2018.

\bibitem{lemeur_liu}
O.~Le~Meur and Z.~Liu, ``Saccadic model of eye movements for free-viewing
  condition,'' \emph{Vision research}, vol. 116, pp. 152--164, 2015.

\bibitem{boccignone2004modelling}
G.~Boccignone and M.~Ferraro, ``Modelling gaze shift as a constrained random
  walk,'' \emph{Physica A: Statistical Mechanics and its Applications}, vol.
  331, no. 1-2, pp. 207--218, 2004.

\bibitem{sam}
M.~Cornia, L.~Baraldi, G.~Serra, and R.~Cucchiara, ``A deep multi-level network
  for saliency p rediction,'' in \emph{Pattern Recognition (ICPR), 2016 23rd
  International Conference on}.\hskip 1em plus 0.5em minus 0.4em\relax IEEE,
  2016, pp. 3488--3493.

\bibitem{kummerer2016deepgaze}
M.~K{\"u}mmerer, T.~S. Wallis, and M.~Bethge, ``Deepgaze ii: Reading fixations
  from deep features trained on object recognition,'' \emph{arXiv preprint
  arXiv:1610.01563}, 2016.

\bibitem{mit-saliency-benchmark}
Z.~Bylinskii, T.~Judd, A.~Borji, L.~Itti, F.~Durand, A.~Oliva, and A.~Torralba,
  ``Mit saliency benchmark.''

\bibitem{koch1984selecting}
C.~Koch and S.~Ullman, ``Selecting one among the many: A simple network
  implementing shifts in selective visual attention.'' MASSACHUSETTS INST OF
  TECH CAMBRIDGE ARTIFICIAL INTELLIGENCE LAB, Tech. Rep., 1984.

\bibitem{coutrot}
A.~Coutrot and N.~Guyader, ``Toward the introduction of auditory information in
  dynamic visual attention models,'' in \emph{Image Analysis for Multimedia
  Interactive Services (WIAMIS), 2013 14th International Workshop on}.\hskip
  1em plus 0.5em minus 0.4em\relax IEEE, 2013, pp. 1--4.

\bibitem{fleiss1971measuring}
J.~L. Fleiss, ``Measuring nominal scale agreement among many raters.''
  \emph{Psychological bulletin}, vol.~76, no.~5, p. 378, 1971.

\end{thebibliography}

\end{document}